\documentclass[10pt,twocolumn,letterpaper]{article}
\usepackage[T1]{fontenc}

\usepackage[final]{cvpr}

\usepackage{amsmath,amssymb,amsfonts}
\usepackage{comment}
\usepackage{wrapfig}
\usepackage{algorithm}
\usepackage{algpseudocode}
\usepackage{hhline}
\usepackage{rotating}
\usepackage{multirow}
\usepackage{CJKutf8}
\usepackage{tabularx}
\usepackage{subcaption}
\usepackage{array}
\usepackage{cellspace}
\usepackage{makecell}
\usepackage{textcomp}
\usepackage{enumitem}
\usepackage{threeparttable}
\usepackage{microtype}
\usepackage{booktabs}
\usepackage{graphicx}
\usepackage{svg}
\PassOptionsToPackage{table}{xcolor}
\usepackage{xcolor}
\usepackage{colortbl}
\usepackage{float}

\usepackage[numbers,sort&compress]{natbib}

\definecolor{cvprblue}{rgb}{0.21,0.49,0.74}
\usepackage[pagebackref,breaklinks,colorlinks,allcolors=cvprblue]{hyperref}

\providecommand{\ccsdesc}[2][]{}
\providecommand{\keywords}[1]{}
\providecommand{\received}[2][]{}
\providecommand{\acmConference}[3][]{}
\providecommand{\acmISBN}[1]{}
\providecommand{\acmDOI}[1]{}
\providecommand{\acmYear}[1]{}
\providecommand{\copyrightyear}[1]{}
\providecommand{\setcopyright}[1]{}
\providecommand{\settopmatter}[1]{}

\def\paperID{0000}
\def\confName{arXiv}
\def\confYear{2026}

\title{\textbf{Adversarial Orthogonal Disentanglement for LVLM Hallucination Mitigation}}

\author{
    Ruoxi Cheng$^{1,2,*}$,\;
    Haoxuan Ma$^{3,*}$,\;
    Zhengfei Hai$^{4}$,\;
    Yiyan Huang$^{5}$,\;\\
    Ranjie Duan$^{2}$,\;
    Tianle Zhang$^{6}$,\;
    Xu Yang$^{4}$,\;
    Ziyi Ye$^{1}$,\;
    Xingjun Ma$^{1,\dag}$ \\
    $^1$Fudan University \quad
    $^2$Tencent \quad
    $^3$Nanjing University \quad
    $^4$Southeast University \\
    $^5$Great Bay University \quad
    $^6$TeleAI, China Telecom
}

\begin{document}

\maketitle

\renewcommand{\thefootnote}{\fnsymbol{footnote}}
\footnotetext[1]{Co-first authors.}
\footnotetext[2]{Corresponding author: \href{mailto:xingjunma@fudan.edu.cn}{xingjunma@fudan.edu.cn}.}
\renewcommand{\thefootnote}{\arabic{footnote}}

\begin{abstract}

Large Vision-Language Models (LVLMs) have revolutionized multimodal understanding. However, a critical barrier to their reliable deployment is hallucination, where generated descriptions conflict with visual facts.
Early hallucination mitigation approaches relying on external knowledge or instruction tuning incur high alignment tax and latency. 
Recent internal interventions also falter: attention-based sparsification creates a logical paradox by pruning via defective weights, while statistical classifiers on raw hidden states overlook the severe feature entanglement between truthful semantics and hallucinatory noise.
To address these limitations, we propose \textbf{A}dversarial \textbf{O}rthogonal \textbf{D}isentanglement (AOD) to mitigate LVLM hallucinations via latent geometric decomposition. 
AOD learns a dedicated hallucination direction using a minimax adversarial objective: a classifier concentrates hallucination signals into the projection, while an adversary actively purges them from the orthogonal residual space via a Gradient Reversal Layer (GRL).
Leveraging this singular direction, we employ a training-free inference strategy—dual-forward-pass contrastive decoding. This intervention effectively suppresses hallucinations while preserving foundational capabilities, enabling robust zero-shot transfer to unseen tasks. 
Extensive experiments on three representative LVLMs across four visual hallucination and four general utility benchmarks demonstrate that AOD significantly outperforms state-of-the-art baselines. Specifically, AOD yields substantial performance gains, raising POPE accuracy by over 6\% on average and providing a 6\% boost on AMBER. AOD also exhibits robust performance on complex utility tasks like MMMU, effectively preserving semantic richness. Our analysis further shows that these hallucination directions transfer robustly across different datasets, suggesting that AOD captures a universal bias rather than dataset-specific artifacts.
Our source code and datasets are available at \href{https://github.com/Hunter-Wrynn/AOD}{https://github.com/Hunter-Wrynn/AOD}.

\end{abstract}

\section{Introduction}

While Large Vision-Language Models (LVLMs) demonstrate profound multimodal capabilities~\citep{li2024vision,cheng2025ecoalign}, their tendency for hallucination—generating textual content inconsistent with actual visual facts—presents a critical risk~\citep{sapkota2026object,xia2025rethinking}. This flaw fundamentally compromises their reliability and safety~\citep{cheng2025inverse,cheng2025pbi}, especially in high-stakes applications such as autonomous driving and medical diagnosis~\citep{zhu2025can}. Therefore, understanding the root mechanisms of these hallucinations and developing robust mitigation strategies remains a paramount research priority.

\begin{figure*}[t]
\centering
\includegraphics[width=\linewidth]{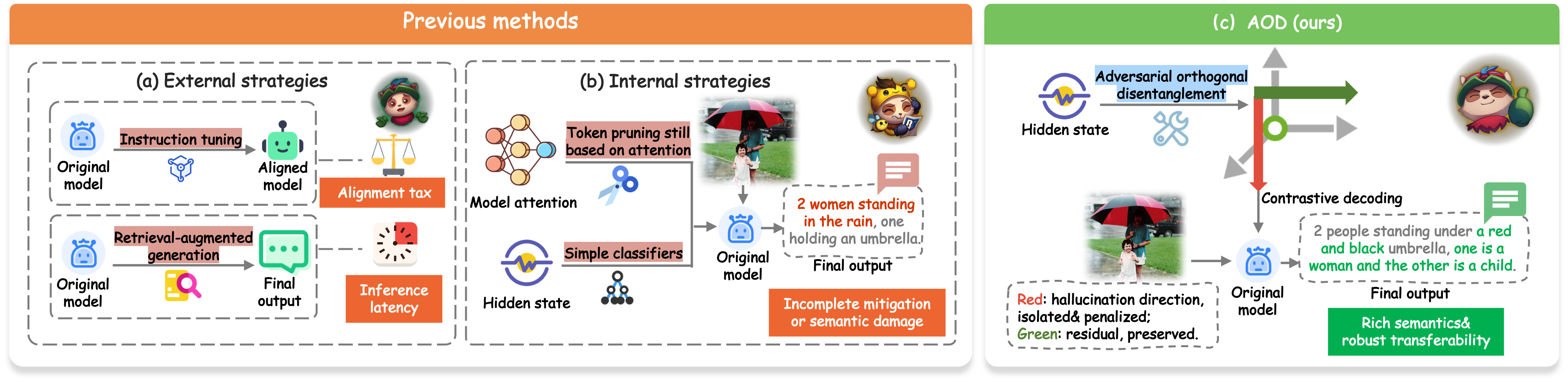}
\caption{\small \textbf{Comparison of hallucination mitigation strategies in LVLMs.} 
\textbf{(a) External strategies} (e.g., instruction tuning, RAG) often suffer from severe alignment taxes or significant inference latency.
\textbf{(b) Internal strategies} rely on paradoxical attention-based pruning or employ simple classifiers that struggle to disentangle truthful semantics from hallucinatory noise, often destroying useful features and yielding vague outputs. 
\textbf{(c) Our AOD framework} leverages adversarial orthogonal disentanglement to effectively separate the hallucination direction from the semantic residual. Through contrastive decoding, AOD specifically penalizes hallucinatory features, achieving robust mitigation while largely preserving semantic richness.
\label{fig:home}}
\end{figure*}

Current hallucination mitigation methods broadly fall into external and internal strategies as shown in \Cref{fig:home}. 
\textbf{(a) External strategies} primarily rely on alignment-based optimization or knowledge retrieval, imposing prohibitive computational overheads. For instance, instruction tuning~\citep{schulman2017proximal} incurs costly training and ``alignment tax'' that degrades foundational reasoning to reduce hallucinations, while RAG~\citep{lewis2020retrieval} introduces severe inference latency, hindering real-time deployment. 
Consequently, focus has shifted to \textbf{(b) Internal strategies} that intervene within the model architecture, yet these also frequently fall short.
Attention-based token pruning~\citep{sun2025prunehal} is fundamentally paradoxical, as it relies on the very defective attention weights that trigger hallucinations to filter them. Alternatively, intervening on hidden state tensors shows promise~\citep{duan2025truthprint}, but existing methods employ simplistic binary classifiers that fail to disentangle truthful visual semantics from hallucinatory noise within the highly coupled latent space. 
Consequently, these crude, brute-force interventions either fail to eradicate hallucinations entirely or inadvertently compromise core semantic features, rendering generated texts vague, conservative, and semantically impoverished~\citep{varela2026entanglement}.

To overcome the limitations of existing internal strategies, particularly the entanglement of factual and hallucinatory features, we propose \textbf{A}dversarial \textbf{O}rthogonal \textbf{D}isentanglement (AOD), which reframes hallucination mitigation as a latent geometric decomposition problem, aiming to enhance the separation of these features within high-dimensional hidden space.
The core motivating insight of AOD is that effective hallucination mitigation necessitates meticulously separating hallucination-prone features from general visual semantics, thereby enabling targeted intervention without eroding foundational capabilities. 
To achieve this geometric separation, AOD optimizes a minimax adversarial objective: a consistency classifier concentrates hallucination-predictive signals along a compact projected direction, while an adversary actively purges these signals from the orthogonal residual space via a Gradient Reversal Layer (GRL)~\citep{10.5555/2946645.2946704}. Leveraging this cleanly isolated direction, we employ a training-free inference strategy: activation-steered contrastive decoding. By shifting the hidden states along this learned vector, we construct both hallucination-amplified and factual-steered representations. This dual-forward-pass mechanism dynamically penalizes hallucinatory logits, ensuring faithful text generation while fully safeguarding the model's rich semantic residual.


Extensive experiments across diverse LVLM architectures demonstrate that AOD significantly outperforms state-of-the-art baselines. AOD effectively suppresses hallucinations—evidenced by average gains of \textbf{+6.4\%} on POPE and \textbf{+6.0\%} on AMBER—while simultaneously enhancing general utility, such as a \textbf{+10.4\%} boost on OCRBench. This shows that AOD mitigates hallucinations without compromising foundational capabilities. Furthermore, the learned hallucination direction exhibits exceptional zero-shot transferability: extracted from a single distribution, it seamlessly generalizes across varying difficulty levels and diverse hallucination typologies (objects, attributes, and relations). 
These findings suggest that AOD captures a universal hallucination bias, establishing a robust, highly efficient framework for mitigating hallucination across diverse vision-language models.

Our main contributions are as follows:

\begin{itemize}
    \item We propose the \textbf{A}dversarial \textbf{O}rthogonal \textbf{D}isentanglement (AOD) framework, framing hallucination mitigation as a latent geometric decomposition. Using a minimax objective, AOD mitigates the entanglement of hallucination signals while largely preserving visual semantics in residual space.

   \item Leveraging this enhanced disentanglement, we employ a training-free contrastive decoding strategy, which precisely penalizes hallucination features during inference to ensure more faithful and grounded visual descriptions.

   \item Extensive evaluations across various models and benchmarks demonstrate AOD's superiority. It achieves state-of-the-art hallucination mitigation while preserving foundational capabilities, providing a highly transferable, training-free solution for reliable deployment across diverse tasks.
\end{itemize}

\begin{figure*}[t]
\centering
\includegraphics[width=\linewidth]{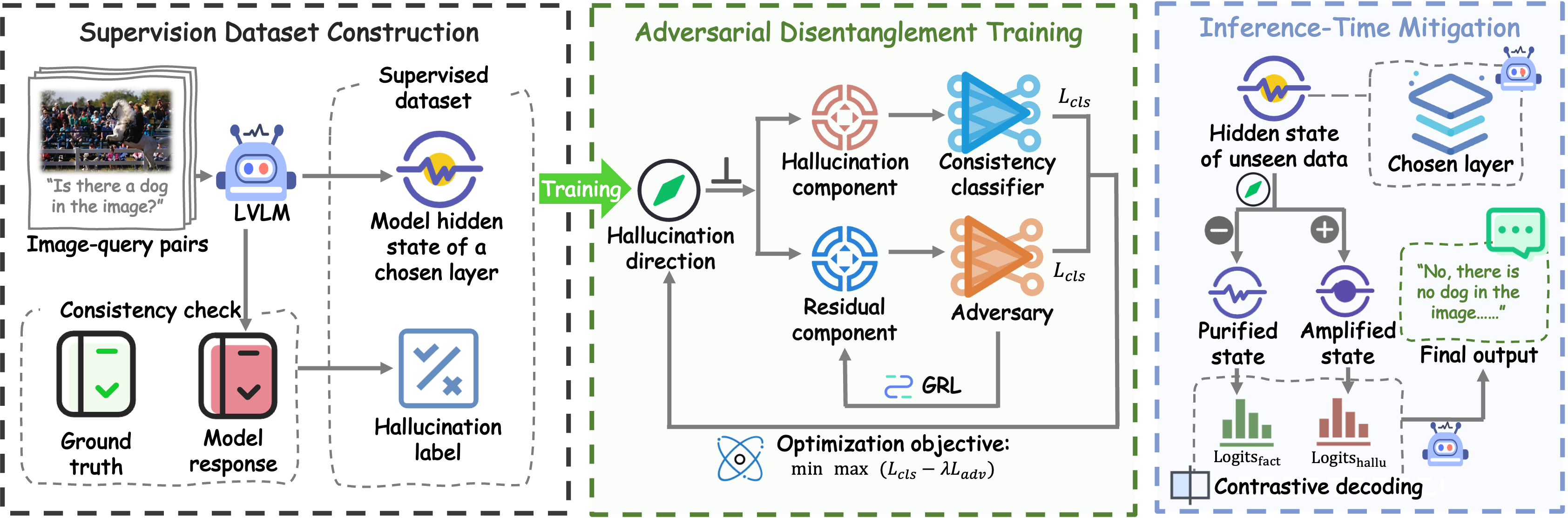}
\caption{\small Overview of the AOD framework. We first extract hidden states from a chosen LVLM layer and assign binary hallucination labels based on the consistency between the model's response and the ground truth. Then we learn a compact hallucination direction that decomposes the hidden state into a projected hallucination component and an orthogonal residual. A minimax objective is employed: a consistency classifier forces hallucination signals to concentrate in the projection, while an adversary with a Gradient Reversal Layer (GRL) actively purges them from the residual to safeguard general semantics. Finally, during inference time, the learned direction enables a training-free, dual-forward-pass contrastive decoding strategy. By deriving a purified state (subtracting the direction) and an amplified state (adding it), AOD dynamically penalizes hallucinatory logits to yield factually grounded final outputs on unseen dataset.
\label{fig:pipeline}}
\end{figure*}

\section{Related Work}

\subsection{Hallucination Mitigation for LVLMs}
The landscape of hallucination mitigation in Large Vision-Language Models (LVLMs) is broadly divided into external augmentations and internal mechanistic interventions. 
External strategies typically utilize curated data or outside knowledge bases, employing techniques like instruction tuning~\citep{kim2025context,leng2024mitigating} and Retrieval-Augmented Generation (RAG)~\citep{ayala2024reducing,park2024mitigating}. While effective, these approaches demand substantial computational overhead or alter foundational model behaviors. 
To circumvent these costs, recent research prioritizes training-free internal interventions. One prominent trajectory employs attention-based sparsification~\citep{zhuang2025vasparse,sun2025prunehal,zhang2026vib,yang2025mitigating} to prune supposedly irrelevant visual tokens. However, these methods depend on the model's inherently flawed attention maps, risking the reinforcement of initial visual-semantic misalignments~\citep{zhao2025mitigating,yangmitigating}. 
Closer to our approach are hidden state interventions. Methods such as ASD~\citep{su-etal-2025-activation} and TruthPrInt~\citep{duan2025truthprint} apply mean activation shifts or leverage classifier probes to rectify hallucinatory states. Yet, by treating hallucinations as simplistic binary shifts, they ignore the complex coupling of factual semantics and noise in latent spaces. Consequently, such blunt interventions either fail to fully eradicate hallucinations or inadvertently compromise fine-grained visual semantics, leading to overly conservative generation.

\subsection{Hallucination Disentanglement}

Disentangled representation learning aims to map complex observations into a structured latent space where distinct generative factors are cleanly separated~\citep{higgins2017betavae,locatello2019challenging}, a paradigm traditionally leveraged for controllable generation~\citep{karras2019style} and bias mitigation~\citep{cadene2019rubi}. Beyond its recent utility in hallucination quantification~\citep{benazha2024measuring,pang2026steering}, disentanglement offers a principled framework for mitigation by isolating the underlying factors of variation that can lead to hallucinatory artifacts~\citep{ma2025watch}. 
Following this principle, several methodologies have emerged to decouple entangled multimodal signals. Prominent approaches include causal interventions designed to sever the confounding link between true visual evidence and over-reliant language priors~\citep{hu2025causal, wang2024causal}. Concurrently, semantic decomposition strategies focus on isolating object-centric representations from complex background noise~\citep{hu2023task, wu2021exemplar}. More recent efforts extend disentanglement to internal model mechanics, such as neutralizing disruptive attention hijackers~\citep{chen2025attention} or utilizing representation engineering to extract linear truthfulness vectors from activations~\citep{zou2023representation}. 
While current disentanglement techniques rely on macroscopic attention maps or heavy structural changes, they fail to address the microscopic entanglement within hidden states. AOD bridges this gap via latent geometric decomposition, surgically isolating hallucination direction while largely preserving uncorrupted semantic residual.

\section{Methodology}

In this section, we present the Adversarial Orthogonal Disentanglement (AOD) framework as shown in \Cref{fig:pipeline}. AOD isolates hallucination-related signals into a compact latent direction while explicitly preserving general visual semantics in the orthogonal subspace. This precise disentanglement enables training-free inference interventions that robustly transfer to unseen tasks.

\subsection{Problem Formulation and Supervision Dataset Construction}
\label{sec:problem_setup}

We study hallucination-related representation disentanglement in the hidden states of an LVLM $\mathcal{M}$. 
Given a dataset $\mathcal{D}=\{(I_i, q_i, a_i^*)\}_{i=1}^m$, each sample consists of an image $I_i$, a question $q_i$ (e.g., ``Is there a [object] in this image?''), and the corresponding ground-truth answer $a_i^*$.
For each sample, we feed the image-question pair $(I_i, q_i)$ into $\mathcal{M}$ and extract the hidden activation vector $\mathbf{x}_{i,l} \in \mathbb{R}^d$ from the $l$-th transformer layer at the final token position\footnote{Our framework is agnostic to the specific token representation; other readout positions or pooling strategies are applicable. We adopt the final token hidden state for its simplicity and consistency across layers.}.

To construct supervision, we assign a binary label $y_i \in \{0,1\}$ according to model response consistency with ground-truth answer:

\[
y_i =
\begin{cases}
1, & \text{if the model prediction matches } a_i^*,\\
0, & \text{otherwise.}
\end{cases}
\]

While this binary label strictly denotes factual consistency, incorrect visually grounded responses in binary LVLM evaluations predominantly correlate with hallucinatory behaviors~\citep{li2023evaluating}. Thus, the learned vector can serve as hallucination-correlated direction.

Accordingly, $y_i=1$ denotes a \emph{consistent} sample and $y_i=0$ denotes an \emph{inconsistent} sample. We use the consistency labels as supervision for learning hallucination-related direction, following the common practice of treating incorrect visually grounded responses as hallucination-associated behaviors in binary LVLM evaluation settings~\citep{su2025activation}. 
This yields a layer-specific training set
\[
\mathcal{D}_l=\{(\mathbf{x}_{i,l}, y_i)\}_{i=1}^m.
\]

\subsection{Orthogonal Decomposition with a Learnable Hallucination Direction}
\label{sec:optimization}

Recent studies suggest that many behavioral properties of large language and vision-language models are linearly encoded in hidden representations~\cite{subramani2022extracting}, and that hallucination-related signals may concentrate in relatively low-dimensional subspaces~\cite{su-etal-2025-activation}. Motivated by these findings, we seek a compact direction that captures the dominant hallucination-related variation while leaving the remaining representation as unaltered as possible.

Specifically, for each layer $l$, we learn a unit vector $\mathbf{v} \in \mathbb{R}^d$ (where $\|\mathbf{v}\|_2=1$) which serves as a hallucination-related direction.

Given a hidden state $\mathbf{x}\in\mathbb{R}^d$, we decompose it through orthogonal projection as
\begin{equation}
    \mathbf{h}_{\text{hallu}} = (\mathbf{x}^\top \mathbf{v})\mathbf{v}, 
    \qquad
    \mathbf{h}_{\text{res}} = \mathbf{x} - (\mathbf{x}^\top \mathbf{v})\mathbf{v}.
\end{equation}

Here, $\mathbf{h}_{\text{hallu}}$ denotes the component of $\mathbf{x}$ along $\mathbf{v}$, while $\mathbf{h}_{\text{res}}$ is the residual component. Since $\|\mathbf{v}\|_2=1$, it trivially holds that $\mathbf{h}_{\text{hallu}}^\top \mathbf{h}_{\text{res}} = 0$, making the decomposition exact and orthogonal by construction.
Notably, orthogonality alone doesn't ensure semantic disentanglement; instead, the distinct roles of these components are induced by the specific supervised objectives.

\subsection{Adversarial Disentanglement Objective}
\label{sec:adversarial_objective}

To encourage the learned direction to capture hallucination-related information, we optimize the decomposition jointly with two discriminators: a consistency classifier $D_C$ and an adversarial residual predictor $D_A$, both parameterized as 3-layer MLPs with parameters $\theta_C$ and $\theta_A$, respectively.

\paragraph{Consistency prediction on the projected component.}
We first require the projected component $\mathbf{h}_{\text{hallu}}$ to be predictive of the consistency label $y$. To this end, the classifier $D_C$ takes $\mathbf{h}_{\text{hallu}}$ as input and is trained with the binary cross-entropy loss

\begin{equation}
\begin{aligned}
    \mathcal{L}_{\text{cls}}(\mathbf{v}, \theta_C) ={}& -\mathbb{E}_{(\mathbf{x}, y) \sim \mathcal{D}_l} \big[ y \log D_C(\mathbf{h}_{\text{hallu}}) \\
    & + (1-y) \log (1 - D_C(\mathbf{h}_{\text{hallu}})) \big].
\end{aligned}
\end{equation}

Minimizing this term encourages the learned direction $\mathbf{v}$ to retain consistency-related predictive cues.

\paragraph{Adversarial suppression on the residual component.}
At the same time, we encourage the residual representation $\mathbf{h}_{\text{res}}$ to be less predictive of $y$. We introduce an adversary $D_A$ that attempts to recover the label from $\mathbf{h}_{\text{res}}$ using:

\begin{equation}
\begin{aligned}
    \mathcal{L}_{\text{adv}}(\mathbf{v}, \theta_A) ={}& -\mathbb{E}_{(\mathbf{x}, y) \sim \mathcal{D}_l} \big[ y \log D_A(\mathbf{h}_{\text{res}}) \\
    & + (1-y) \log (1 - D_A(\mathbf{h}_{\text{res}})) \big].
\end{aligned}
\end{equation}

The adversary is optimized to minimize this loss, while $\mathbf{v}$ is optimized to make residual-based prediction difficult. This yields the following minimax objective:
\begin{equation}
\min_{\mathbf{v},\,\theta_C}\max_{\theta_A}
\Big(
\mathcal{L}_{\text{cls}}(\mathbf{v}, \theta_C)
-
\lambda \mathcal{L}_{\text{adv}}(\mathbf{v}, \theta_A)
\Big),
\end{equation}
where $\lambda>0$ balances the strength of residual invariance.

Intuitively, this objective encourages a concentration of label-predictive information into the projected component $\mathbf{h}_{\text{hallu}}$, while discouraging the residual component $\mathbf{h}_{\text{res}}$ from carrying easily decodable consistency signals. In practice, this does not guarantee full statistical independence of $\mathbf{h}_{\text{res}}$ and $y$, but it provides a direct optimization bias toward a cleaner separation of hallucination-related and residual features.

\paragraph{Practical optimization.}
We implement the above minimax training with a Gradient Reversal Layer (GRL)~\citep{10.5555/2946645.2946704} inserted between $\mathbf{h}_{\text{res}}$ and $D_A$. During the forward pass, the GRL acts as the identity function. During backpropagation, it multiplies the gradient flowing from $D_A$ to $\mathbf{v}$ by $-\lambda$, effectively achieving the adversarial maximization step for $\mathbf{v}$ seamlessly during standard gradient descent, thereby encouraging $\mathbf{v}$ to move toward directions for which the residual representation becomes less informative for predicting $y$. Unless otherwise stated, we enforce the unit-norm constraint on $\mathbf{v}$ by $\ell_2$ normalization after each parameter update.

\subsection{Inference-Time Mitigation}
\label{sec:inference}

Once the hallucination direction $\mathbf{v}^*$ is learned via our adversarial framework, we leverage established contrastive decoding strategies~\citep{leng2024mitigating,su2025activation} to steer the LVLM toward factual generation during inference. This approach utilizes the isolated direction $\mathbf{v}^*$ to dynamically amplify factual signals while suppressing hallucination patterns in the final output distribution.

Given a hidden state $\mathbf{z}$ from the model's chosen layer during generation, we first compute two steered hidden states by injecting the learned direction $\mathbf{v}^*$ with opposite polarities. Specifically, we define a factual-steered state $\mathbf{z}^{+}$ by subtracting the hallucination component, and a hallucination-steered state $\mathbf{z}^{-}$ by adding it:
\begin{equation}
\begin{aligned}
\label{eq:steered_states}
\mathbf{z}^{+} &= \mathbf{z} - \gamma (\mathbf{z}^\top \mathbf{v}^*)\mathbf{v}^*, \\
\mathbf{z}^{-} &= \mathbf{z} + \gamma (\mathbf{z}^\top \mathbf{v}^*)\mathbf{v}^*,
\end{aligned}
\end{equation}
where $\gamma \ge 0$ is the steering strength. Since $\mathbf{v}^*$ is a unit vector, $\gamma$ functions as an absolute scaling factor. In practice, $\gamma$ is normalized relative to the average norm of the projected components to ensure stable intervention.

Next, these steered hidden states are passed through the model's final language head (classifier) to obtain their corresponding unnormalized log probabilities (logits). Let $\text{logit}_{\pi^+}$ and $\text{logit}_{\pi^-}$ denote the logits computed from the factual-steered state $\mathbf{z}^{+}$ and the hallucination-steered state $\mathbf{z}^{-}$, respectively. Following standard contrastive decoding practice, the final logits for token prediction are computed as:
\begin{equation}
\label{eq:contrastive_logits}
\text{logit}_{\text{final}} = (1 + \beta) \text{logit}_{\pi^+} - \beta \text{logit}_{\pi^-},
\end{equation}
where $\beta \ge 0$ is a contrastive weight. This mechanism effectively amplifies the probabilistic difference between factual and hallucinatory behaviors defined by our learned direction.

To maintain generation coherence and prevent severe deviations from the language distribution, we also adopt the Adaptive Plausibility Constraint (APC) proposed in VCD~\citep{leng2024mitigating}, which restricts the contrastive penalty to only high-probability tokens. This entire inference intervention is training-free, relying solely on the robust direction $\mathbf{v}^*$ extracted from our supervised disentanglement phase.

\begin{table*}[t]
\centering
\caption{\small Comparison of AOD and baseline methods across visual hallucination and general utility dimensions. Green values indicate absolute improvements over the base performance. \textbf{Note:} For CHAIR, lower values are better, thus a decrease is highlighted in green.}
\label{tab:ComparisonTable}

\definecolor{gaincolor}{HTML}{38761D} 
\definecolor{losscolor}{HTML}{CC0000} 
\definecolor{neutralcolor}{gray}{0.6} 
\definecolor{rowgray}{HTML}{F5F5F5}
\definecolor{aodblue}{HTML}{E8F0FE}

\newcommand{\gain}[1]{\textcolor{gaincolor}{\small$\uparrow$#1}}
\newcommand{\loss}[1]{\textcolor{losscolor}{\small$\downarrow$#1}}
\newcommand{\chairgain}[1]{\textcolor{gaincolor}{\small$\downarrow$#1}}
\newcommand{\chairloss}[1]{\textcolor{losscolor}{\small$\uparrow$+#1}}
\newcommand{\nogain}[1]{\textcolor{neutralcolor}{\small$\uparrow$0.0}}

\setlength{\tabcolsep}{3.5pt} 
\renewcommand{\arraystretch}{1.2}

\resizebox{\textwidth}{!}{%
\begin{tabular}{l l cccc | cccc}
\toprule
\makecell[c]{\multirow{2}{*}{\textbf{Model}}} &
\makecell[c]{\multirow{2}{*}{\textbf{Method}}}
& \multicolumn{4}{c|}{\textbf{Visual Hallucination}}
& \multicolumn{4}{c}{\textbf{General Utility}} \\
\cmidrule(lr){3-6} \cmidrule(lr){7-10}
& 
& \textbf{POPE} ($\uparrow$) & \textbf{CHAIR} ($\downarrow$) & \textbf{HallusionBench} ($\uparrow$) & \textbf{AMBER} ($\uparrow$)
& \textbf{OCRBench-v2} ($\uparrow$) & \textbf{RealWorldQA} ($\uparrow$) & \textbf{MMStar} ($\uparrow$) & \textbf{MMMU} ($\uparrow$) \\
\midrule

& Base                                  & 84.5\,\nogain{} & 15.3\,\nogain{} & 58.8\,\nogain{} & 78.6\,\nogain{} & 16.8\,\nogain{} & 55.3\,\nogain{} & 32.6\,\nogain{} & 34.5\,\nogain{} \\
\rowcolor{rowgray} & VCD~\citep{leng2024mitigating}        & 87.6\,\gain{3.1} & 14.3\,\chairgain{1.0} & 61.2\,\gain{2.4} & 82.3\,\gain{3.7} & 19.4\,\gain{2.6} & 54.2\,\loss{1.1} & 34.1\,\gain{1.5} & 34.0\,\loss{0.5} \\
& ASD~\citep{su2025activation}          & 88.3\,\gain{3.8} & 13.8\,\chairgain{1.5} & 62.5\,\gain{3.7} & 84.1\,\gain{5.5} & 23.6\,\gain{6.8} & 56.9\,\gain{1.6} & 35.8\,\gain{3.2} & 36.7\,\gain{2.2} \\
\rowcolor{rowgray} & TruthPrInt~\citep{duan2025truthprint} & 89.8\,\gain{5.3} & 14.1\,\chairgain{1.2} & 63.1\,\gain{4.3} & 85.2\,\gain{6.6} & 25.1\,\gain{8.3} & 57.1\,\gain{1.8} & 35.2\,\gain{2.6} & 37.4\,\gain{2.9} \\
& VASparse~\citep{zhuang2025vasparse}   & 87.8\,\gain{3.3} & 14.9\,\chairgain{0.4} & 61.8\,\gain{3.0} & 83.5\,\gain{4.9} & 17.2\,\gain{0.4} & 55.6\,\gain{0.3} & 33.5\,\gain{0.9} & 33.5\,\loss{1.0} \\
\rowcolor{rowgray} & PruneHal~\citep{sun2025prunehal}      & -                & 15.1\,\chairgain{0.2} & 62.7\,\gain{3.9} & 84.6\,\gain{6.0} & 20.2\,\gain{3.4} & 56.1\,\gain{0.8} & 35.1\,\gain{2.5} & 33.6\,\loss{0.9} \\

\rowcolor{aodblue}\multirow{-7}{*}{\textbf{LLaVA-1.5-7B}} & \textbf{AOD (Ours)} & \textbf{90.9\,\gain{6.4}} & \textbf{13.3\,\chairgain{2.0}} & \textbf{64.3\,\gain{5.5}} & \textbf{86.7\,\gain{8.1}} & \textbf{27.2\,\gain{10.4}}& \textbf{57.3\,\gain{2.0}} & \textbf{36.3\,\gain{3.7}} & \textbf{38.3\,\gain{3.8}} \\
\midrule

& Base                                  & 89.0\,\nogain{} & 8.4\,\nogain{} & 63.7\,\nogain{} & 80.6\,\nogain{} & 46.7\,\nogain{} & 68.9\,\nogain{} & 58.9\,\nogain{} & 51.9\,\nogain{} \\
\rowcolor{rowgray} & VCD~\citep{leng2024mitigating}        & 90.1\,\gain{1.1} & 7.9\,\chairgain{0.5} & 64.5\,\gain{0.8} & 81.5\,\gain{0.9} & 47.8\,\gain{1.1} & 66.4\,\loss{2.5} & 59.6\,\gain{0.7} & 49.8\,\loss{2.1} \\
& ASD~\citep{su2025activation}          & 91.5\,\gain{2.5} & 7.2\,\chairgain{1.2} & 65.8\,\gain{2.1} & 83.0\,\gain{2.4} & 49.2\,\gain{2.5} & 70.3\,\gain{1.4} & 60.7\,\gain{1.8} & 54.0\,\gain{2.1} \\
\rowcolor{rowgray} & TruthPrInt~\citep{duan2025truthprint} & 92.3\,\gain{3.3} & 7.5\,\chairgain{0.9} & 66.2\,\gain{2.5} & 83.9\,\gain{3.3} & 50.1\,\gain{3.4} & 70.8\,\gain{1.9} & 61.1\,\gain{2.2} & 54.9\,\gain{3.0} \\
& VASparse~\citep{zhuang2025vasparse}   & 90.8\,\gain{1.8} & 7.8\,\chairgain{0.6} & 65.1\,\gain{1.4} & 82.1\,\gain{1.5} & 48.5\,\gain{1.8} & 67.5\,\loss{1.4} & 60.1\,\gain{1.2} & 50.2\,\loss{1.7} \\

\rowcolor{rowgray} & PruneHal~\citep{sun2025prunehal}      & - & 6.8\,\chairgain{1.6} & 67.0\,\gain{3.3} & 84.5\,\gain{3.9} & 48.9\,\gain{2.2} & 68.2\,\loss{0.7} & 57.8\,\loss{1.1} & 50.7\,\loss{1.2} \\

\rowcolor{aodblue}\multirow{-7}{*}{\textbf{Qwen2.5-VL-7B}} & \textbf{AOD (Ours)} & \textbf{94.6\,\gain{5.6}} & \textbf{5.0\,\chairgain{3.4}} & \textbf{69.2\,\gain{5.5}} & \textbf{86.8\,\gain{6.2}} & \textbf{52.4\,\gain{5.7}} & \textbf{72.2\,\gain{3.3}} & \textbf{63.0\,\gain{4.1}} & \textbf{57.6\,\gain{5.7}} \\
\midrule

& Base                                  & 84.1\,\nogain{} & 6.9\,\nogain{} & 60.7\,\nogain{} & 91.3\,\nogain{} & 49.0\,\nogain{} & 63.4\,\nogain{} & 55.8\,\nogain{} & 65.0\,\nogain{} \\
\rowcolor{rowgray} & VCD~\citep{leng2024mitigating}        & 86.5\,\gain{2.4} & 6.6\,\chairgain{0.3} & 61.9\,\gain{1.2} & 92.6\,\gain{1.3} & 49.1\,\gain{0.1} & 63.0\,\loss{0.4} & 54.6\,\loss{1.2} & 66.1\,\gain{1.1} \\
& ASD~\citep{su2025activation}          & 85.4\,\gain{1.3} & 6.7\,\chairgain{0.2} & 63.1\,\gain{2.4} & 93.4\,\gain{2.1} & 51.5\,\gain{2.5} & 64.8\,\gain{1.4} & 56.1\,\gain{0.3} & 66.5\,\gain{1.5} \\
\rowcolor{rowgray} & TruthPrInt~\citep{duan2025truthprint} & 89.0\,\gain{4.9} & 6.8\,\chairgain{0.1} & 63.4\,\gain{2.7} & 92.8\,\gain{1.5} & 52.9\,\gain{3.9} & 65.1\,\gain{1.7} & 56.8\,\gain{1.0} & 68.0\,\gain{3.0} \\
& VASparse~\citep{zhuang2025vasparse}   & 85.7\,\gain{1.6} & 6.3\,\chairgain{0.6} & 61.2\,\gain{0.5} & 92.1\,\gain{0.8} & 48.7\,\loss{0.3} & 63.8\,\gain{0.4} & 56.0\,\gain{0.2} & 65.8\,\gain{0.8} \\
\rowcolor{rowgray} & PruneHal~\citep{sun2025prunehal}      & -                & 6.0\,\chairgain{0.9} & 65.1\,\gain{4.4} & 94.0\,\gain{2.7} & 52.1\,\gain{3.1} & 67.0\,\gain{3.6} & 59.0\,\gain{3.2} & 68.9\,\gain{3.9} \\
\rowcolor{aodblue}\multirow{-7}{*}{\textbf{InternVL3-8B}} & \textbf{AOD (Ours)} & \textbf{91.4\,\gain{7.3}} & \textbf{5.6\,\chairgain{1.3}} & \textbf{66.5\,\gain{5.8}} & \textbf{95.0\,\gain{3.7}} & \textbf{54.4\,\gain{5.4}} & \textbf{68.1\,\gain{4.7}} & \textbf{60.2\,\gain{4.4}} & \textbf{69.1\,\gain{4.1}} \\
\bottomrule
\end{tabular}
}
\end{table*}

\section{Evaluation}
\label{sec:evaluation}

We demonstrate the effectiveness of AOD through extensive experiments on various models and benchmarks.

\subsection{Experimental Setup}

\paragraph{Models.}
We evaluate on three open-source LVLMs, specifically LLaVA-1.5-7B~\citep{liu2024improved}, Qwen2.5-VL-7B ~\citep{wang2024qwen2} and InternVL3-8B ~\citep{zhu2025internvl3}.

\paragraph{Baselines.} 
We compare \textbf{AOD} with a range of hallucination mitigation methods. 
\textbf{Base} applies raw prompting.
\textbf{VCD} ~\citep{leng2024mitigating} mitigates hallucinations by contrasting output distributions derived from original and distorted visual inputs.
\textbf{ASD} ~\citep{su2025activation} corrects hallucinations based on mean-difference directional patterns in intermediate activations, serving as a representative linear subspace baseline.
\textbf{TruthPrInt} ~\citep{duan2025truthprint} steers the generation process by applying truthful-guided inference-time interventions within learned latent subspaces.
\textbf{VASparse} ~\citep{zhuang2025vasparse} efficiently preserves visual context and recalibrates attention via visual-aware token sparsification and sparse-based contrastive decoding.
\textbf{PruneHal} ~\citep{sun2025prunehal} utilizes adaptive KV cache pruning to strengthen the model's focus on critical visual information.

\paragraph{Benchmarks.}
We evaluate our method across two primary dimensions: hallucination mitigation and general utility. To assess \textbf{visual hallucination}, we employ POPE~\citep{li2023evaluating} based on the MSCOCO dataset~\citep{leng2024mitigating}, including the three splits (\textit{adversarial}, \textit{random} and \textit{popular}). We further measure hallucination and reasoning robustness using MSCOCO CHAIR~\citep{rohrbach2018object} (here we test CHAIR$_{\text{S}}$, sentence-level), HallusionBench~\citep{guan2024hallusionbench}, and AMBER~\citep{wang2023amber}. To verify \textbf{general utility}, ensuring our intervention preserves fundamental capabilities, we evaluate LVLM performance on OCRBench-v2~\citep{Liu_2024,fu2024ocrbenchv2improvedbenchmark}, RealWorldQA~\citep{realworldqa2024}, MMStar~\cite{chen2024we} and MMMU~\citep{yue2023mmmu}. To ensure a fair and standardized comparison, we report scores following the official settings of all benchmarks.
More details are shown in \Cref{appendix_benchmark}.

\paragraph{Datasets and evaluation protocols.} 
To rigorously evaluate our orthogonal disentanglement, we apply two distinct protocols. For hallucination benchmarks, we adopt an \textit{in-domain evaluation}: each dataset is randomly partitioned into an 80\% split for extracting the task-specific hallucination direction $\mathbf{v}$, and a 20\% unseen split for testing. Conversely, for general utility benchmarks where consistency probes are not easily defined, we directly apply the hallucination direction extracted from POPE. This evaluates whether AOD preserves, rather than relearns, foundational capabilities—validating our method’s robust zero-shot transferability without task-specific re-extraction. More details are shown in \Cref{appendix_protocol}.

\paragraph{Implementation details.}

All experiments are conducted on eight NVIDIA A100 (80GB) GPUs. For results in \Cref{tab:ComparisonTable}, we apply the optimal intervention layers and hyperparameter configurations empirically identified through our ablation studies (\cref{sec:ablation}). To ensure reliability, all reported scores represent the average of four independent trials. More details are shown in \Cref{appendix_implementation}.

\subsection{Main Results}
\label{sec:results}

As shown in \Cref{tab:ComparisonTable}, AOD consistently establishes a new state-of-the-art across all evaluated models, achieving remarkable improvements in both hallucination mitigation and general utility. By isolating hallucination signals into an orthogonal subspace, AOD yields substantial gains on the POPE benchmark: \textbf{+6.4\%} for LLaVA-1.5 and \textbf{+7.3\%} for InternVL3-8B. It also reduces the CHAIR hallucination rate by \textbf{3.4 points} on Qwen2.5-VL. Consistent improvements on HallusionBench and AMBER further demonstrate AOD's efficacy across diverse model scales.
Significantly, AOD avoids the common pitfall where pursuing hallucination reduction inadvertently degrades foundational capabilities. While baselines like VCD and VASparse often sacrifice general utility for safety (e.g., VCD incurs a \textbf{-2.5\%} drop on RealWorldQA; VASparse drops \textbf{-1.0\%} on MMMU), AOD's precise geometric decomposition excises artifacts without collateral damage to the visual-semantic manifold. Beyond merely preserving performance, AOD frequently enhances it, delivering striking boosts such as \textbf{+10.4\%} on OCRBench-v2 and \textbf{+5.7\%} on the challenging MMMU. This dual gain in safety and utility demonstrates the robustness of our orthogonal disentanglement and its seamless transferability to complex, open-ended tasks.

\begin{figure*}[t]
    \centering
    \def\svgwidth{\linewidth} 
    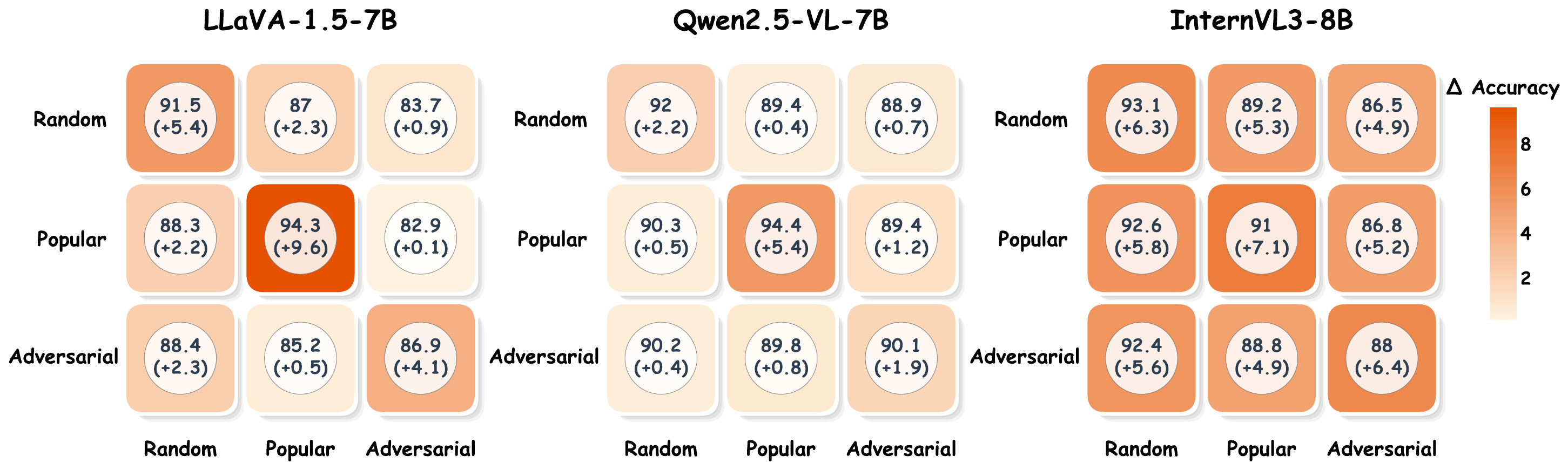 
    \caption{\small \textbf{Cross-difficulty transferability on POPE splits.} Heatmaps illustrating the zero-shot accuracy improvement ($\Delta$) when applying a hallucination direction extracted from one specific difficulty split to others \textbf{Base} indicates original LVLM performance without intervention.}
    \label{fig:heatmap2}
\end{figure*}

\begin{figure*}[t]
    \centering
    \def\svgwidth{\linewidth} 
    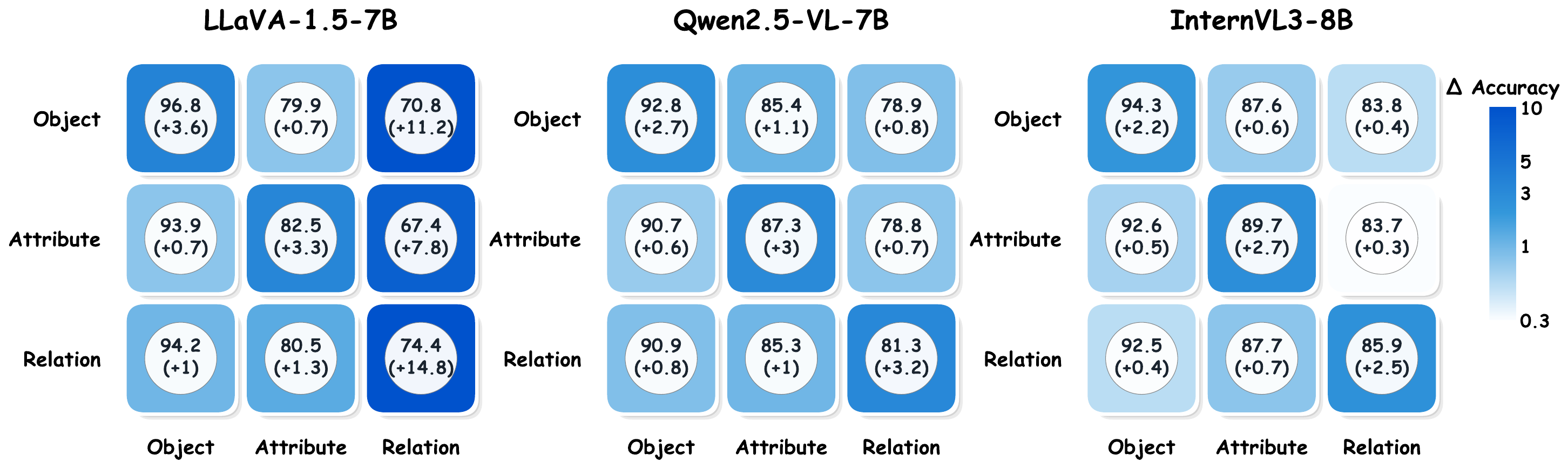 
    \caption{\small \textbf{Cross-typology transferability on AMBER categories.} Heatmaps illustrating the zero-shot accuracy improvement ($\Delta$) when applying a hallucination direction  extracted from one specific hallucination type to others. \textbf{Base} indicates original LVLM performance without intervention.}
    \label{fig:heatmap1}
\end{figure*}

\begin{figure*}[t]
\centering
\includegraphics[width=0.98\linewidth]{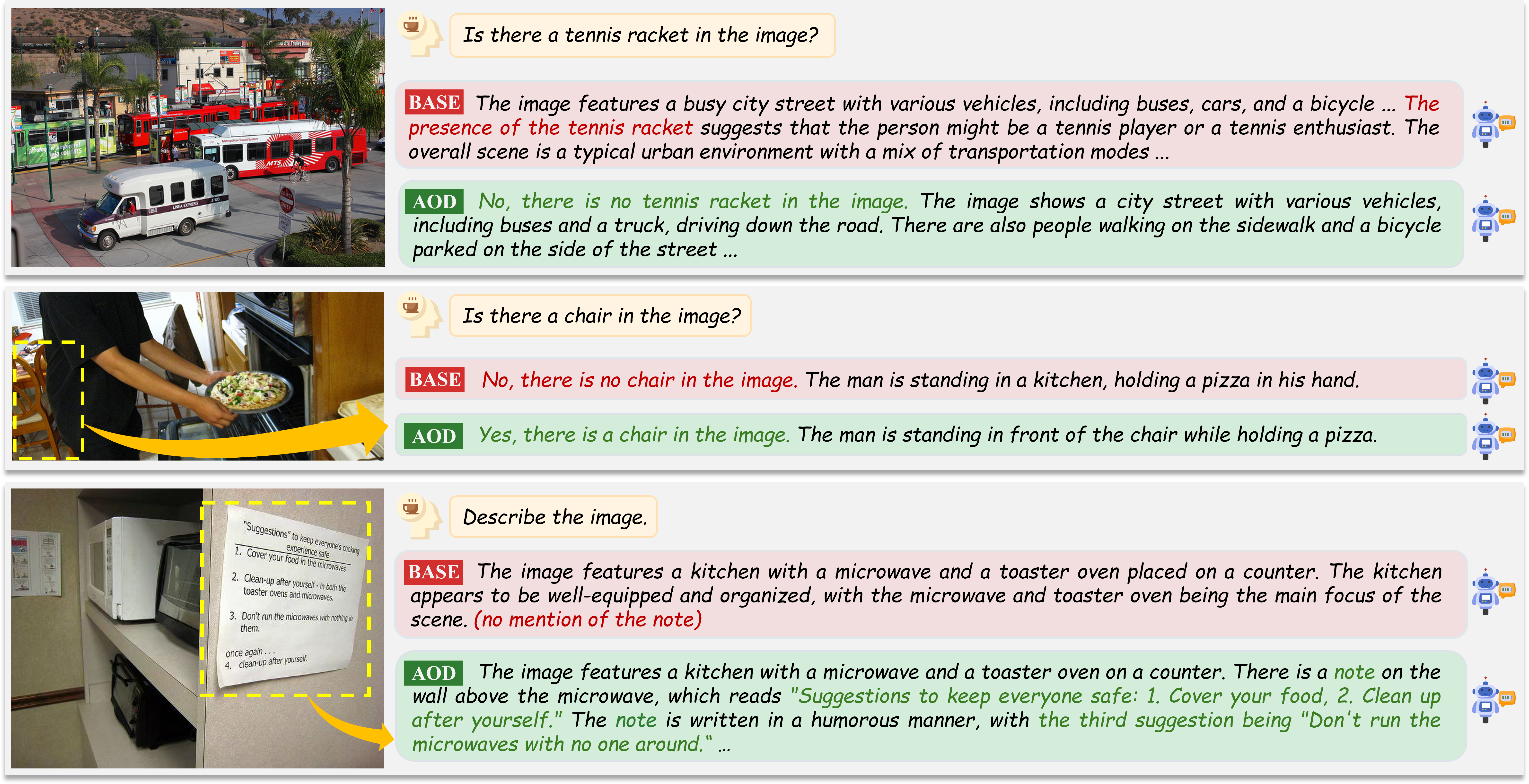}
\caption{\textbf{Qualitative cases of AOD interventions on LLaVA-1.5-7B.} Red text highlights hallucinations or missed visual evidence by the Base model, while green text denotes AOD's factual corrections and precise grounding.}
\label{fig:case_study}
\end{figure*}

\subsection{Analysis}
\label{sec:analysis}

\paragraph{Cross-difficulty transferability.} 
We first investigate whether a hallucination direction $\mathbf{v}$ learned from one specific data distribution generalizes to others with different difficulty levels. Using the POPE benchmark, we extract $\mathbf{v}$ from one split (Random, Popular, or Adversarial) and apply it to unseen splits. As shown in \Cref{fig:heatmap2}, while in-domain interventions yield the highest gains (the diagonal), AOD exhibits robust cross-distribution transferability. Notably, directions extracted from the most challenging \textit{Adversarial} split—which targets complex co-occurrence biases—consistently improve performance on the simpler \textit{Random} and \textit{Popular} splits across all models (e.g., up to +6.4\% on InternVL3). This suggests that AOD captures fundamental cognitive biases rather than merely overfitting to specific dataset artifacts, demonstrating strong cross-domain generalization.

\paragraph{Cross-typology transferability.} 
Multimodal hallucinations manifest not only as fabricated objects but also as incorrect attributes (e.g., wrong colors) or false relations (e.g., spatial misplacements). To determine whether these diverse failure modes share the same latent subspace, we cross-evaluate AOD on the AMBER benchmark's Object, Attribute, and Relation subsets. As detailed in \Cref{fig:heatmap1}, applying a specific intervention vector (e.g., $\mathbf{v}_{obj}$) yields the most significant improvements on its corresponding task, while generally exerting a marginal impact on misaligned tasks. This localized effectiveness demonstrates that different hallucination typologies are geometrically distinct within the latent space. Consequently, AOD functions as a highly decoupled intervention, rectifying specific failure modes without causing cross-semantic interference.

\begin{figure}[t]
\centering
\includegraphics[width=\linewidth]{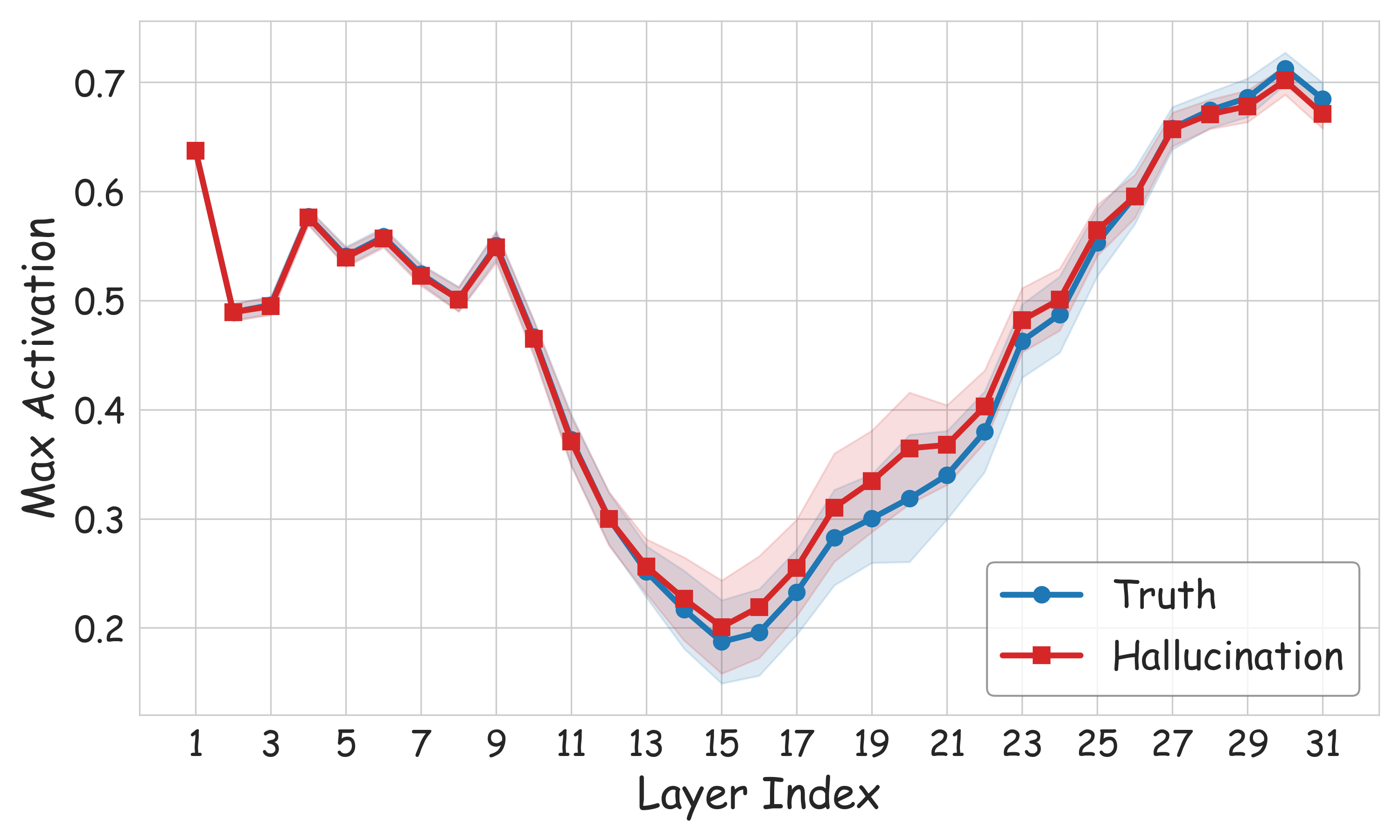}
\caption{\textbf{Layer-wise max activation.} Evolution of normalized hidden state max activations across 31 layers of LLaVA-1.5-7B on AMBER. Solid lines and shaded areas denote mean and standard deviation for truth vs. hallucination.}
\label{fig:significance}
\end{figure}


\begin{figure*}[t]
\centering
    \begin{subfigure}{0.2\textwidth}
        \def\svgwidth{\linewidth}
        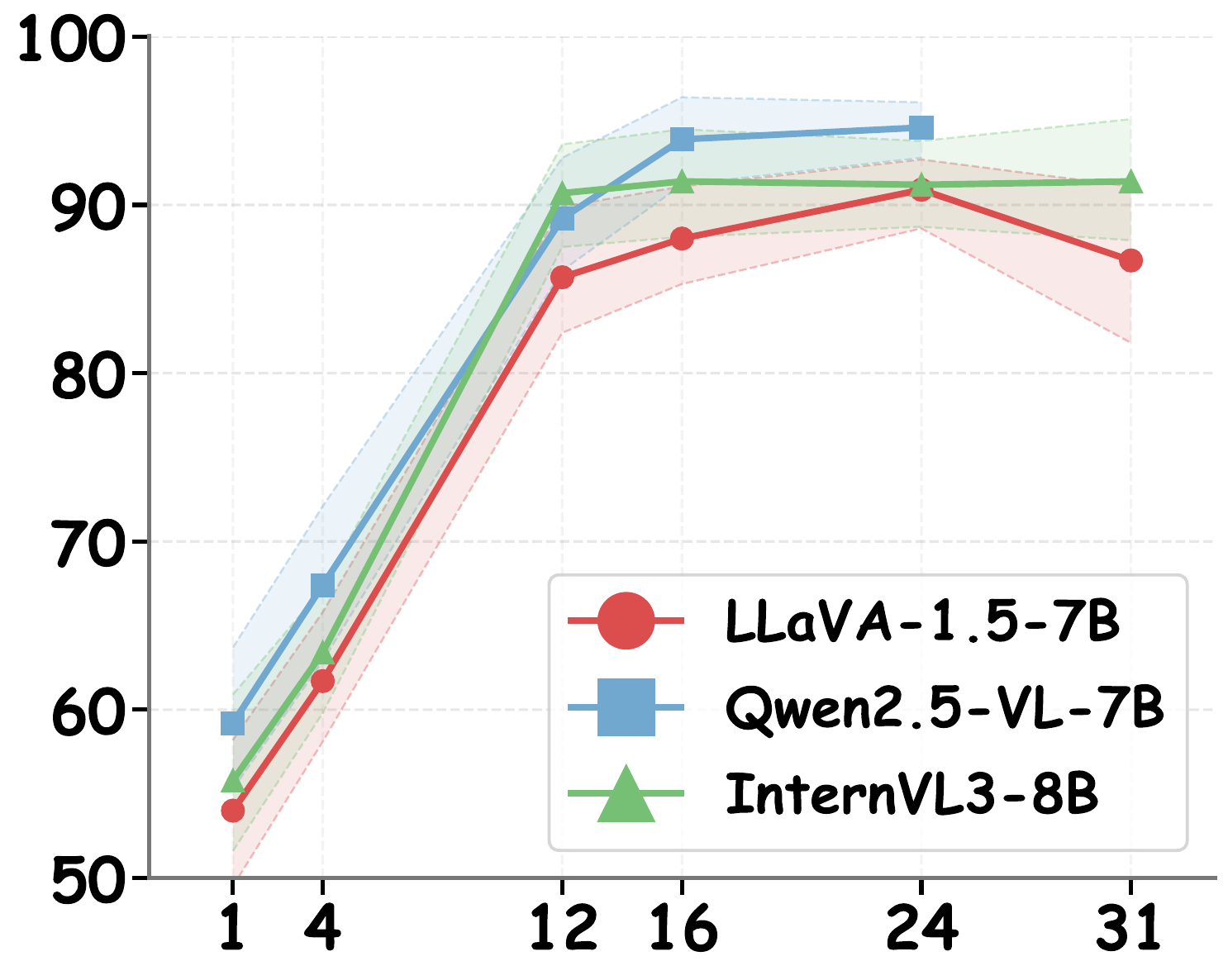
        \caption{Intervention Layer}
        \label{fig:layer}
    \end{subfigure}\hfill
    \begin{subfigure}{0.2\textwidth}
        \def\svgwidth{\linewidth}
        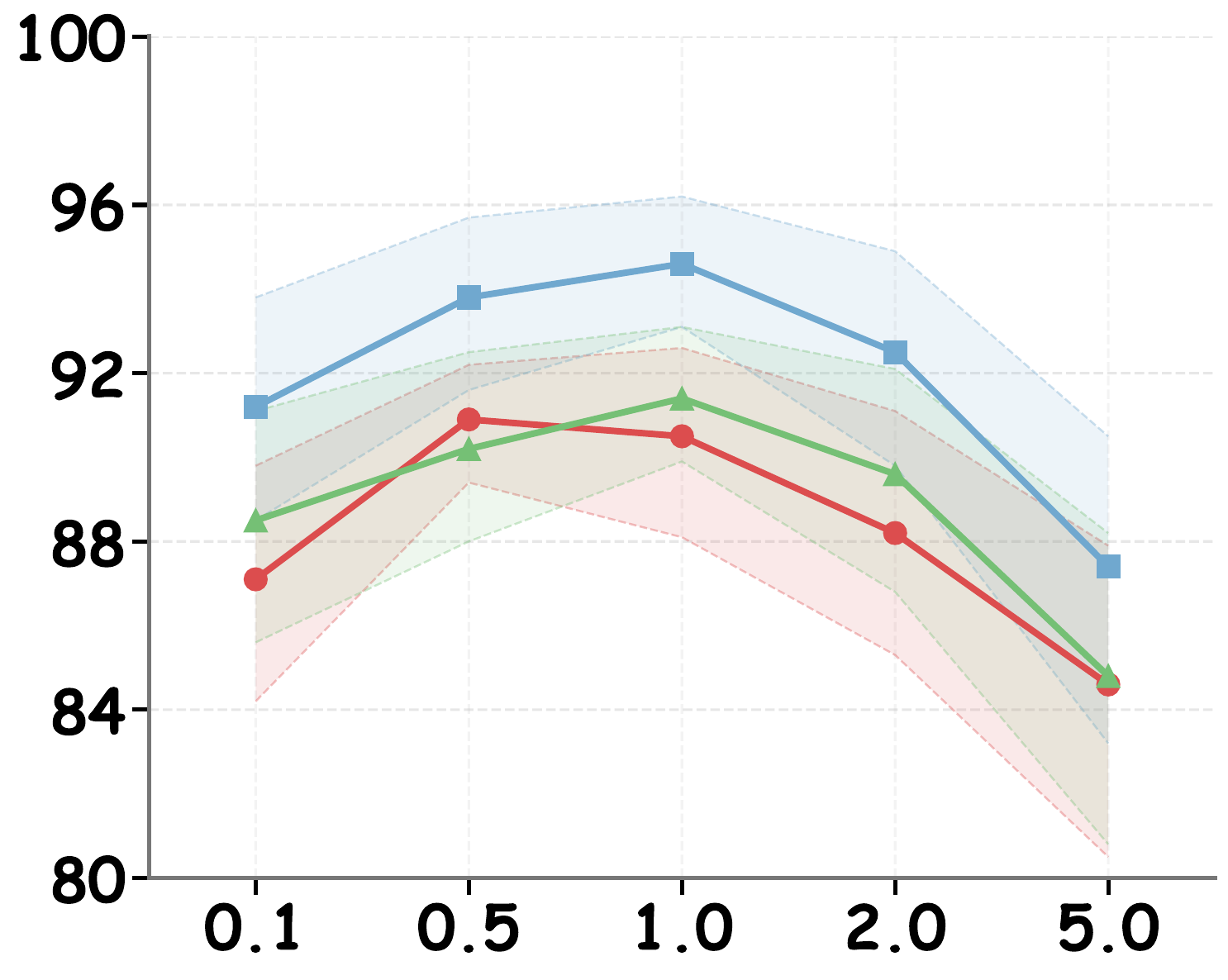
        \caption{Steering Strength ($\gamma$)}
        \label{fig:beta}
    \end{subfigure}\hfill
    \begin{subfigure}{0.2\textwidth}
        \def\svgwidth{\linewidth}
        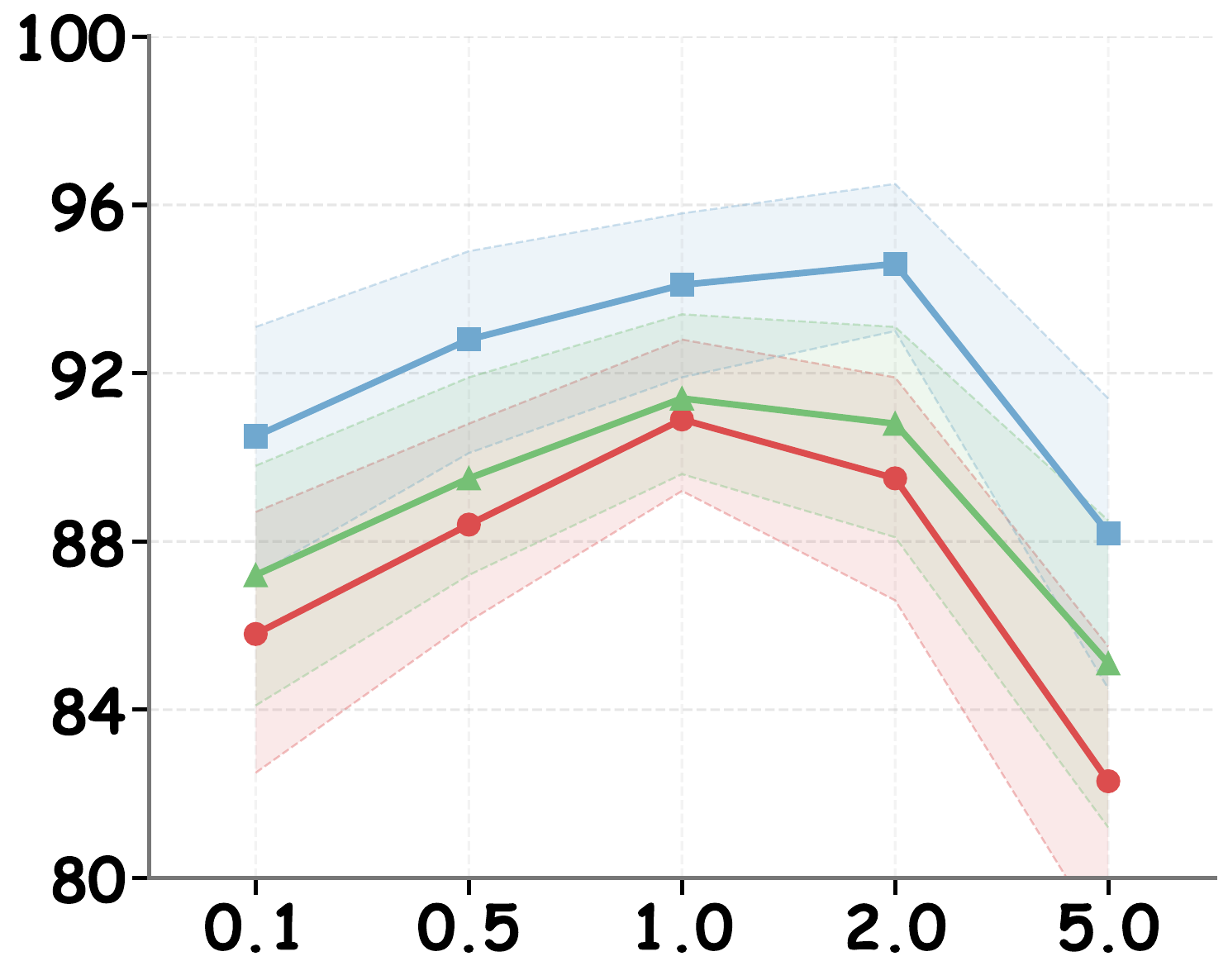
        \caption{Contrastive Weight ($\beta$)}
        \label{fig:gamma}
    \end{subfigure}\hfill
    \begin{subfigure}{0.2\textwidth}
        \def\svgwidth{\linewidth}
        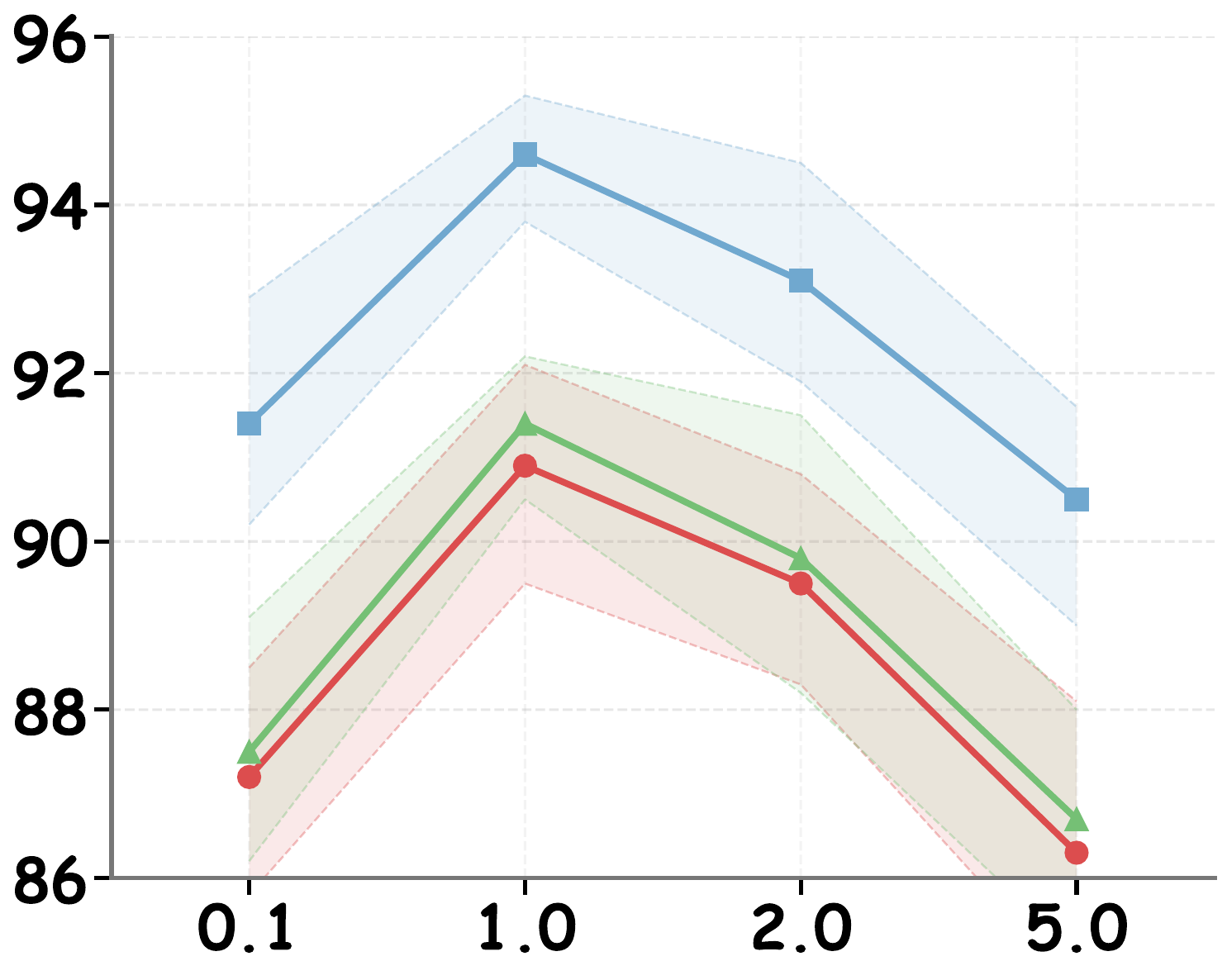
        \caption{Adversarial Weight ($\lambda$)}
        \label{fig:lambda}
    \end{subfigure}\hfill
    \begin{subfigure}{0.2\textwidth}
        \def\svgwidth{\linewidth}
        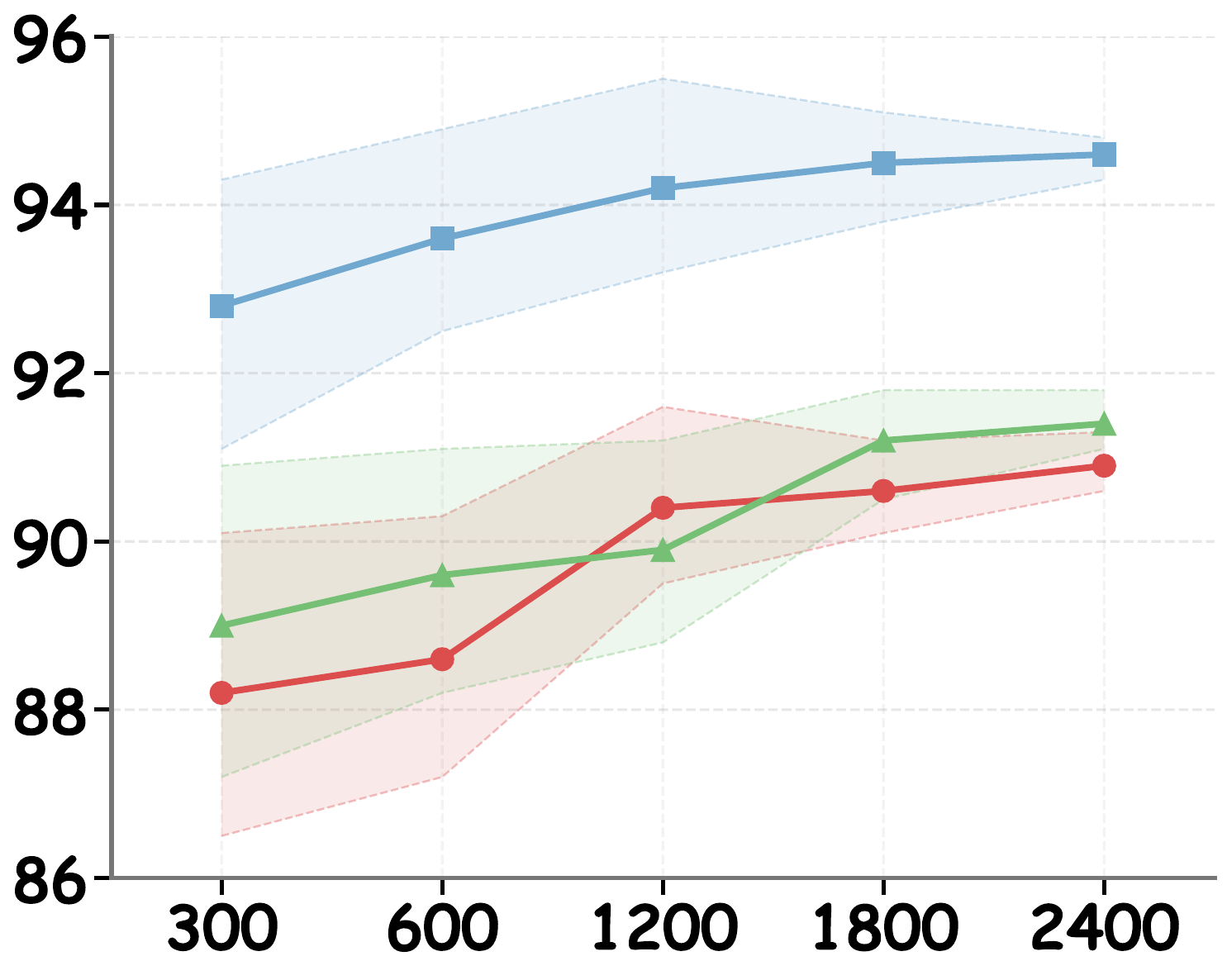
        \caption{Training Dataset Size}
        \label{fig:dataset}
    \end{subfigure}
\caption{Ablation studies on POPE benchmark across three LVLMs: (a) intervention layer, (b) steering strength $\gamma$, (c) contrastive weight $\beta$, (d) adversarial weight $\lambda$, and (e) training dataset size.}
\label{fig:ablation_main}
\end{figure*}

\paragraph{Layer-wise analysis and selection.}
Transformer layers transition from low-level perception to high-level reasoning. To identify the optimal stage for intervention, we analyze activation dynamics and intervention efficacy across all 31 layers of LLaVA-1.5-7B. 
Following \citet{zou2023representation}, who identified outlier maximum activations as indicators of specific latent behavioral signals, we track the max activation values of $\ell_2$-normalized hidden states on the AMBER dataset.
As shown in \Cref{fig:significance}, \textbf{Truth} and \textbf{Hallucination} activation patterns remain entangled in early layers but diverge significantly beyond Layer 15, where hallucination signals become distinctly prominent. This suggests that hallucination patterns magnify during semantic integration rather than early perception. 
This trend is corroborated by direct intervention experiments on POPE (\Cref{fig:layer}), where efficacy remains marginal in shallow layers but peaks robustly in the middle-to-late stages. We hypothesize that the final layers become overly entangled with rigid, token-specific vocabulary projection, whereas the middle-to-late stages represent a ``sweet spot'': a stage where multimodal misalignment is fully manifested yet remains sufficiently decoupled from output tokenization, enabling AOD to precisely intercept  hallucination-inducing signals.

\vspace{-3mm}
\paragraph{Qualitative case study.} 
To intuitively illustrate the robust intervention mechanism of AOD, \Cref{fig:case_study} presents three representative generation scenarios comparing base model with our approach:
\textbf{(1) Mitigating false positives (Top):} Driven by spurious visual correlations or strong language priors, the base model hallucinates a non-existent ``tennis racket.'' By subtracting the isolated hallucination direction, AOD decisively eradicates this fabricated object, re-anchoring model to actual visual evidence for a factually grounded description.
\textbf{(2) Recovering false negatives (Middle):} The base model completely misses the partially obscured chair in the background. Crucially, AOD successfully recovers it, which demonstrates our orthogonal disentanglement does not merely induce a naive, conservative ``no-bias'' (which artificially inflates precision at the cost of recall). Instead, AOD genuinely calibrates representations to reflect the true visual state.
\textbf{(3) Enhancing fine-grained grounding (Bottom):} While base model ignores the wall note, AOD activates fine-grained features to accurately transcribe the dense text. This aligns with our +10.4\% gain on OCRBench-v2, demonstrating that suppressing latent hallucinatory noise can improve visual grounding. These improvements across OCR and complex reasoning tasks suggest that AOD captures a fundamental cognitive bias: the model’s over-reliance on language priors. By penalizing this prior-heavy direction, AOD forces the LVLM to prioritize actual visual evidence, mitigating the language-shortcut errors prevalent in dense reasoning tasks.

\subsection{Ablation Study}
\label{sec:ablation}

We perform comprehensive ablation studies to analyze AOD’s mechanics. Unless noted, all experiments are conducted on the POPE dataset. More details are shown in \Cref{appendix_hyperparameter}.

\begin{table}[t]
\centering
\caption{\small Ablation study of different inference strategies. We report performance on key hallucination and utility benchmarks. The cost is presented as a relative multiplier compared to the Base model. \textbf{AOD w/o CD} denotes the ablated version using only single-forward direct latent intervention.}
\label{tab:ablation_inference_strategies}

\definecolor{gaincolor}{HTML}{38761D}      
\definecolor{losscolor}{HTML}{CC0000}      
\definecolor{neutralcolor}{gray}{0.6}     
\definecolor{rowgray}{HTML}{F5F5F5}
\definecolor{aodblue}{HTML}{E8F0FE}

\newcommand{\gain}[1]{\textcolor{gaincolor}{\small$\uparrow$#1}}
\newcommand{\costincrease}[1]{\textcolor{losscolor}{\small$\uparrow$#1}} 
\newcommand{\nogain}[1]{\textcolor{neutralcolor}{\small$-$\phantom{0.0}}} 

\setlength{\tabcolsep}{4pt}  
\renewcommand{\arraystretch}{1.1} 
\small

\begin{tabular}{l cc c c}
\toprule
\multirow{2}{*}{\textbf{Method}}
& \multicolumn{2}{c}{\textbf{Hallucination}}
& \textbf{Utility}
& \textbf{Cost} \\
\cmidrule(lr){2-3} \cmidrule(lr){4-4} \cmidrule(lr){5-5}
& \textbf{POPE} & \textbf{AMBER}
& \textbf{MMStar}
& \textbf{Run time} \\
\midrule

\multicolumn{5}{l}{\textbf{LLaVA-1.5-7B}} \\
\midrule
Base & 84.5\,\nogain{} & 78.6\,\nogain{} & 32.6\,\nogain{} & 1.0\,\nogain{} \\
\rowcolor{rowgray} AOD w/o CD & 89.0\,\gain{4.5} & 84.5\,\gain{5.9} & 35.1\,\gain{2.5} & 1.1\,\costincrease{0.1} \\
\rowcolor{aodblue} \textbf{AOD (Ours)} & \textbf{90.9}\,\gain{6.4} & \textbf{86.7}\,\gain{8.1} & \textbf{36.3}\,\gain{3.7} & 2.3\,\costincrease{1.3} \\
\midrule

\multicolumn{5}{l}{\textbf{Qwen2.5-VL-7B}} \\
\midrule
Base & 89.0\,\nogain{} & 80.6\,\nogain{} & 58.9\,\nogain{} & 1.0\,\nogain{} \\
\rowcolor{rowgray} AOD w/o CD & 92.5\,\gain{3.5} & 84.0\,\gain{3.4} & 61.4\,\gain{2.5} & 1.1\,\costincrease{0.1} \\
\rowcolor{aodblue} \textbf{AOD (Ours)} & \textbf{94.6}\,\gain{5.6} & \textbf{86.8}\,\gain{6.2} & \textbf{63.0}\,\gain{4.1} & 2.3\,\costincrease{1.3} \\
\midrule

\multicolumn{5}{l}{\textbf{InternVL3-8B}} \\
\midrule
Base & 84.1\,\nogain{} & 91.3\,\nogain{} & 55.8\,\nogain{} & 1.0\,\nogain{} \\
\rowcolor{rowgray} AOD w/o CD & 88.6\,\gain{4.5} & 93.5\,\gain{2.2} & 57.8\,\gain{2.0} & 1.1\,\costincrease{0.1} \\
\rowcolor{aodblue} \textbf{AOD (Ours)} & \textbf{91.4}\,\gain{7.3} & \textbf{95.0}\,\gain{3.7} & \textbf{60.2}\,\gain{4.4} & 2.2\,\costincrease{1.2} \\
\bottomrule
\end{tabular}
\end{table}


\vspace{-3mm}
\paragraph{Intervention layer.} 
\Cref{fig:layer} shows a consistent trend across all models: POPE accuracy increases as intervention moves to deeper layers. LLaVA-1.5 and Qwen2.5-VL peak at Layer 24 (90.9\% and 94.6\%), while InternVL3 remains robust across Layers 16–31 ($>91\%$). This demonstrates that multimodal hallucination stems from high-level semantic misalignment. Targeting middle-to-late layers allows AOD to intercept hallucination signals after semantic integration but before vocabulary projection.

\vspace{-3mm}
\paragraph{Hyperparameters $\gamma$ and $\beta$.} 
We evaluate steering strength $\gamma$ and contrastive weight $\beta$ on POPE as shown in \Cref{fig:beta,fig:gamma}. For $\gamma$, low values (e.g., 0.1) fail to diverge states, while excessive strength (e.g., 5.0) nduces severe negative transfer (over-correction), where valid visual semantics are aggressively penalized along with the noise, ultimately causing semantic collapse and degraded reasoning. $\gamma \in [1.0, 2.0]$ is optimal. Similarly, $\beta \in [0.5, 1.0]$ balances hallucination penalty and fluency, whereas $\beta > 2.0$ harms reasoning. AOD remains robust across these ranges, ensuring stable deployment.

\vspace{-1mm}
\paragraph{Adversarial weight $\lambda$.} 
The parameter $\lambda$ controls gradient reversal strength to purify $\mathbf{h}_{\text{res}}$ of hallucination signals. As shown in \Cref{fig:lambda}, $\lambda=0$ causes the framework to degenerate into a standard linear probe classifier without orthogonal disentanglement; without adversarial penalty, entangled features lead to suboptimal mitigation and utility degradation. An optimal $\lambda=1.0$ enforces rigorous geometric disentanglement, achieving the best trade-off between hallucination suppression and capability preservation.

\vspace{-1mm}
\paragraph{Training dataset size.} 
AOD is highly data-efficient as shown in \Cref{fig:dataset}. It identifies stable hallucination directions with only 300--600 samples (e.g., 93.6\% for Qwen2.5-VL at 600 samples). Performance plateaus beyond 1,800 samples, where LLaVA-1.5 and InternVL3 reach 90.6\% and 91.2\%. This efficiency enables rapid, low-cost deployment in resource-constrained scenarios without compromising model integrity.

\vspace{-1mm}
\paragraph{Inference strategies.} 
Comparing AOD against \textbf{AOD w/o CD} (single-forward latent intervention) reveals a clear trade-off as shown in \Cref{tab:ablation_inference_strategies}. \textbf{AOD w/o CD} is a lightweight alternative, mitigating hallucinations (+4.5 POPE on LLaVA-1.5) with negligible overhead ($1.1\times$ cost). The full \textbf{AOD} yields the highest gains—improving InternVL3's POPE by an additional 2.8 points over the ablated version—by dynamically amplifying factual signals at a $2.2\times$ latency cost.
While AOD incurs higher inference latency than standard prompting, it avoids the prohibitive data-curation and model-wide retraining costs inherent in instruction tuning, and circumvents the cumbersome retrieval pipelines required by RAG.

\section{Conclusion}
\label{sec:conclusion}
In this paper, we propose Adversarial Orthogonal Disentanglement (AOD) to address the deep entanglement between truthful visual semantics and hallucinatory noise in LVLMs. By reframing hallucination mitigation as a latent geometric decomposition problem, AOD utilizes a minimax objective to surgically isolate the dominant hallucination-correlated direction while safeguarding foundational representations within the orthogonal residual space. Leveraging this geometric separation, our activation-steered contrastive decoding dynamically penalizes hallucinatory patterns during inference, effectively suppressing errors without requiring backbone retraining. Extensive experiments demonstrate that AOD establishes a new state-of-the-art across diverse benchmarks. By neutralizing the model's over-reliance on language priors, it achieves substantial gains in both hallucination suppression and complex utility tasks. Furthermore, the learned hallucination directions exhibit robust zero-shot transferability across distinct hallucination typologies and out-of-domain evaluation setups, providing a highly transferable, training-free mechanism for reliable multimodal deployment.

\clearpage
\newpage

\bibliographystyle{abbrvnat}
\bibliography{main}

\clearpage
\newpage
\appendix

\section{Appendix}
\label{sec:appendix}

\subsection{Benchmark Details}
\label{appendix_benchmark}

\paragraph{POPE}
POPE is a polling-style object hallucination benchmark built on MSCOCO images. Each example asks a binary question about the presence or absence of an object. Our local snapshot contains the standard \textit{adversarial}, \textit{random}, and \textit{popular} splits, with 3,000 questions per split. POPE serves as the primary benchmark for learning $\mathbf{v}^*$ and for the main hallucination-mitigation ablations.

\paragraph{CHAIR}
We use the sentence-level CHAIR metric on MSCOCO-based image captioning outputs to quantify object hallucination in open-ended generation. Lower CHAIR scores indicate fewer hallucinated object mentions. In the main paper, CHAIR complements POPE by measuring whether the intervention remains effective when the output is no longer restricted to binary answers.

\paragraph{HallusionBench.}
HallusionBench evaluates multimodal hallucination and reasoning failures under carefully designed visual conditions. The local snapshot used in this repository contains 1,129 examples. The data include both visually dependent and visually supplementary settings, together with metadata such as category, subcategory, and visual-input type. In our evaluation, these samples are converted into binary yes/no judgments for consistent scoring.

\paragraph{AMBER}
AMBER is a multimodal hallucination benchmark containing both generative and discriminative evaluation protocols. Our appendix experiments focus on the discriminative subsets because they provide clean binary supervision for extracting and testing hallucination directions. The local snapshot contains 4,924 existence examples, 7,628 attribute examples, and 1,664 relation examples. Table 8 uses these three typologies to study whether a direction learned on one hallucination type transfers to the others.

\paragraph{OCRBench-v2.}
OCRBench measures text perception and OCR-centric reasoning in multimodal models. The repository contains both OCRBench v1 (1,000 test examples) and OCRBench v2 (10,000 test examples). The main paper reports OCR-oriented utility preservation under the official benchmark metric, while the released code supports both versions for compatibility across model families.

\paragraph{RealWorldQA}
RealWorldQA focuses on visual understanding in real-world scenes. The local snapshot used in this repository contains 765 test examples. We use this benchmark as an additional utility-oriented evaluation to verify that the learned direction transfers beyond hallucination-specific datasets.

\paragraph{MMStar.}
MMStar is a vision-indispensable multiple-choice benchmark designed to reduce language-only shortcuts and dataset leakage. The local snapshot contains 1,500 validation examples spanning 6 core capability groups and 18 fine-grained axes. It is particularly useful in our study because it provides a strong test of whether hallucination mitigation damages broader multimodal competence.

\paragraph{MMMU}
MMMU evaluates broad multimodal understanding across many academic disciplines. The prepared local snapshot in this repository contains a 900-example validation subset spanning 30 subject configurations. We use MMMU as a high-level utility benchmark to check whether the learned hallucination direction preserves knowledge-intensive reasoning.

\subsection{Benchmark-Specific Evaluation Protocols}
\label{appendix_protocol}

\paragraph{Binary hallucination benchmarks (POPE, HallusionBench, AMBER)}
For POPE, HallusionBench, and the discriminative subsets of AMBER, the primary evaluation metric is accuracy:
\[
\mathrm{Acc} = \frac{1}{N}\sum_{i=1}^N \mathbb{I}\big[\hat{a}_i = a_i\big].
\]
To reduce formatting variance and ambiguity, we prompt the models to produce concise ``yes'' or ``no'' outputs. If a model fails to produce a normalized valid answer, the sample is strictly counted as incorrect.

\paragraph{CHAIR}
For open-ended captioning, we quantify object hallucinations using the CHAIR metric based on MSCOCO. Let $C$ be the total number of generated captions, and $C_{\mathrm{hall}} \subseteq C$ be the subset of captions containing at least one hallucinated MSCOCO object. We compute the sentence-level hallucination rate ($\mathrm{CHAIR}_S$) as:
\[
\mathrm{CHAIR}_S = \frac{|C_{\mathrm{hall}}|}{|C|}.
\]
Lower values indicate fewer hallucinations. During evaluation, object mentions are canonicalized using the standard CHAIR synonym list alongside a lightweight singularization heuristic.

\paragraph{OCRBench-v2.}
We evaluate visual text localization and reasoning capabilities using OCRBench-v2, a comprehensive text-centric benchmark containing 10,000 human-verified image-question pairs. To ensure rigorous evaluation, the benchmark employs task-specific scoring mechanisms. Multiple-choice and regression-style numeric questions are parsed using predefined rules, while open-ended OCR tasks use exact or near-exact string matching after normalization. This normalization process removes trivial formatting discrepancies, such as repeated whitespaces, punctuation variants, and standard answer wrappers.

\paragraph{RealWorldQA}
RealWorldQA is designed to assess the real-world spatial understanding and physical reasoning capabilities of multimodal models. It comprises over 700 high-resolution, authentic images sourced from everyday environments and driving scenarios, each paired with a verifiable question. We evaluate models under the default zero-shot setting, measuring performance based on the accuracy of the extracted correct choice or short answer.

\paragraph{MMStar.}
MMStar evaluates core vision-language capabilities using multiple-choice questions with up to five candidate options. To accurately measure performance and reduce the variance caused by verbose explanations frequently generated by LVLMs, we restrict the generation length and normalize the model's prediction by extracting the first valid option letter. The reported metric is the overall multiple-choice accuracy.

\paragraph{MMMU}
For the Massive Multi-discipline Multimodal Understanding (MMMU) benchmark, scoring strictly depends on the question's answer type. Multiple-choice questions are evaluated by extracting the predicted option letter or normalizing the option text. Numeric questions are graded after number normalization, while free-form text questions rely on normalized exact string matching. The final benchmark score is reported as the overall accuracy across all evaluated disciplines.

\subsection{Details}
\label{appendix_implementation}

\paragraph{Backbone LVLMs.}
Our main experiments use three representative backbones: LLaVA-1.5-7B, Qwen2.5-VL-7B, and InternVL3-8B. In our released codebase, these correspond to the Hugging Face checkpoints \path{llava-hf/llava-1.5-7b-hf}, \path{Qwen/Qwen2.5-VL-7B-Instruct}, and \path{OpenGVLab/InternVL3-8B-hf}. All models are loaded in \texttt{bfloat16} with automatic device mapping.

\paragraph{Experimental pipeline overview.}
To ensure reproducibility, we summarize the AOD pipeline, which decouples hallucination mitigation from the backbone LVLM's parameters into four stages:
\begin{enumerate}
    \item \textbf{Base Inference \& Labeling:} We process input pairs $(x_i, q_i)$ through the frozen base LVLM to obtain predictions $\hat{a}_i$. Consistency labels $y_i \in \{0, 1\}$ are generated by comparing $\hat{a}_i$ with ground-truth answers $a_i$. For POPE-style experiments, prompts explicitly force single-word answers to ensure unambiguous labels.
    \item \textbf{Hidden State Extraction:} We extract hidden activations $z_i \in \mathbb{R}^d$ at the intervention layer $\ell$ corresponding to the last generated token position (i.e., \texttt{token\_index=-1}). For all main experiments, we set $\ell=24$.
    \item \textbf{Adversarial Training:} We learn the hallucination-correlated direction $\mathbf{v}^*$ using the extracted pairs $\{(z_i, y_i)\}_{i=1}^N$ via the minimax objective.
    \item \textbf{Inference Intervention:} Once trained, $\mathbf{v}^*$ acts as a plug-and-play steering vector. It can be dynamically applied during inference via contrastive decoding or reused across unseen datasets without backbone fine-tuning.
\end{enumerate}

\paragraph{Optimization and training parameters.}
Given a hidden representation $z$ and a unit-norm direction $v$, AOD decomposes the representation into a hallucination-aligned component $h_{\mathrm{hallu}} = (z^\top v)v$ and an orthogonal residual $h_{\mathrm{res}} = z - (z^\top v)v$. We train $v$ together with two three-layer MLP probes: a classifier $D_C$ operating on the projected component and an adversary $D_A$ operating on the residual component through a Gradient Reversal Layer. Using GRL, the optimization can be written as:
\[
\min_{v,\theta_C,\theta_A}
\sum_i \mathcal{L}_{\mathrm{BCE}}\big(D_C(h_{\mathrm{hallu},i}), y_i\big)
+ \mathcal{L}_{\mathrm{BCE}}\!\big(D_A(\mathrm{GRL}_\lambda(h_{\mathrm{res},i})), y_i\big),
\]
where the forward pass of $\mathrm{GRL}_\lambda$ is the identity and the backward pass multiplies the gradient by $-\lambda$. In the implementation, $v$ is renormalized to unit norm after every update. The main sweeps use the AdamW optimizer with a learning rate of $10^{-3}$, batch size $256$, hidden size $512$, validation ratio $0.2$, and a random seed of $42$. All reported models use 5 training epochs. Unless otherwise stated, the adversarial weight is fixed to $\lambda=1.0$.

\paragraph{Inference-time hyperparameters.}
The paper notation uses $\gamma$ for the intervention strength and $\beta$ for the contrastive weight. (In the released code, $\gamma$ is mapped to the command-line flag \texttt{aod\_alpha}). For direct intervention, applying $\tilde{z} = z - \gamma (z^\top v^*) v^*$ with $\gamma=1.0$ exactly removes the projection of $z$ onto $v^*$, leaving only the orthogonal residual $h_{\mathrm{res}}$.

For contrastive decoding, we define the bidirectional steered states $z^{\pm} = z \mp \gamma (z^\top v^*) v^*$. Let $\pi_{+}$ and $\pi_{-}$ denote the next-token distributions induced by $z^+$ and $z^-$. The final contrastive logit is computed as $\mathrm{logit}_{\mathrm{final}} = (1+\beta)\,\mathrm{logit}_{+} - \beta\,\mathrm{logit}_{-}$. 
Following VCD, we additionally apply an Adaptive Plausibility Constraint (APC) to avoid deviating too far from the base distribution. If $p_{+}$ is the softmax of $\mathrm{logit}_{+}$ and $\tau=\alpha_{\mathrm{APC}}\max_j p_{+}(j)$, only tokens satisfying $p_{+}(j)\ge \tau$ are allowed to use the contrastive logit; all others fall back to the positive branch logits. Unless a table explicitly sweeps these hyperparameters, we use the best validation-selected configuration for each model.

\subsection{Supplementary Observations on Hyperparameter Behavior}
\label{appendix_hyperparameter}

\paragraph{Effect of the adversarial weight $\lambda$.}
The parameter $\lambda$ determines how aggressively the residual component is purified through adversarial pressure. In our experiments, we observe a distinct trade-off between geometric disentanglement and intervention efficacy. As shown in the ablation studies, setting $\lambda=0$ (or using a very weak penalty like $\lambda=0.1$) causes the framework to degenerate into a standard linear probe. Under this setting, while the model maintains high utility scores (e.g., MMStar remains around $59.2$ for Qwen2.5-VL), the entangled nature of the features leads to sub-optimal hallucination suppression. The isolated direction fails to cleanly capture the hallucination bias, resulting in significantly lower POPE and AMBER scores compared to the optimal setting. Conversely, an optimal $\lambda=1.0$ enforces rigorous geometric disentanglement, effectively purging hallucinatory noise from the residual space. This achieves the best overall trade-off—maximizing hallucination mitigation (e.g., driving peak POPE scores) while incurring only a negligible fluctuation in foundational capabilities.

\paragraph{Effect of the intervention strength $\gamma$.}
The direct latent intervention strategy requires careful tuning of the steering strength $\gamma$ to balance factual calibration with semantic coherence. Across LLaVA-1.5, Qwen2.5-VL, and InternVL3, performance remains highly stable and optimal near $\gamma \approx 1.0$, which mathematically corresponds to the exact removal of the projected hallucination component. However, pushing the strength to extreme values aggressively penalizes latent features, inducing severe negative transfer. For instance, when forcing $\gamma = 5.0$, we observe that valid visual semantics are over-corrected, causing the model's performance to degrade sharply on both hallucination benchmarks and open-ended reasoning tasks. This semantic collapse demonstrates that excessive intervention destroys the underlying manifold, validating $\gamma \in [1.0, 2.0]$ as the principled operational range.

\paragraph{Effect of the contrastive weight $\beta$.}
The contrastive parameter $\beta$ determines the intensity of the penalty applied to the hallucination-steered negative branch during decoding. Moderate contrastive weighting ($\beta \in [0.5, 1.0]$) effectively sharpens the factual distribution, yielding consistent gains across all three evaluated LVLMs. However, the final output distribution is highly sensitive to an overly amplified negative branch. When $\beta$ is increased beyond $2.0$ (e.g., $\beta=5.0$), the decoding process is overwhelmed by the penalty, leading to severe disruptions in the language model's inherent fluency and reasoning coherence. This results in noticeable performance drops across complex utility benchmarks, demonstrating that the negative branch should act strictly as a targeted regularizer rather than a dominant decoding signal.

\paragraph{Cost-performance trade-off of inference strategies.}
The single-forward direct intervention strategy (\textbf{AOD w/o CD}) computationally mirrors the base model ($1.1\times$ latency cost), making it highly suitable for real-time applications. In contrast, activation-steered contrastive decoding (\textbf{AOD}) requires dual forward passes per decoding step, increasing the inference latency by roughly a factor of two ($2.2\times \sim 2.3\times$). As evidenced in \Cref{tab:ablation_inference_strategies}, direct intervention successfully recovers the majority of the anti-hallucination gains. Contrastive decoding pushes the boundary further, dynamically amplifying factual signals to achieve maximum precision when additional computation is permissible. Therefore, these two strategies represent flexible operating points on the same efficiency--performance Pareto frontier, allowing users to tailor AOD to specific deployment constraints.

\end{document}